\documentclass[letterpaper, 10 pt, conference]{ieeeconf}
\usepackage{PackageHeading}
\usepackage[absolute,overlay]{textpos}

\usepackage{booktabs}
\usepackage{siunitx}
\usepackage{caption}

\makeatletter
\@ifundefined{labelindent}{}{\let\labelindent\relax}
\makeatother
\usepackage[shortlabels]{enumitem} % or your enumitem options

\usepackage{enumitem}
\usepackage{xcolor,colortbl} % Add this to the preamble
\usepackage{array}    % For specifying column format

\usepackage{tikz}
\usetikzlibrary{positioning}
\usepackage{svg}

\usepackage{makecell} % Required for custom line breaks within cells

\title{\Large \bf
Learning Dolly-In Filming From Demonstration Using a Ground-Based Robot
}

\author{Philip Lorimer$^{1}$, Alan Hunter$^{2}$, and Wenbin Li$^{1}$% <-this % stops a space
\thanks{This work was supported by EPSRC Centre for Digital Entertainment with the grant number EP/L016540/1.}% <-this % stops a space
\thanks{$^{1}$Department of Computer Science,
        University of Bath, UK, \{\protect\url{pall20, w.li}\}\protect\url{@bath.ac.uk}}
\thanks{
        $^{2}$Department of Mechanical Engineering,
        University of Bath, UK, \protect\url{A.J.Hunter@bath.ac.uk}}%
}

\begin{document}

% \begin{textblock*}{12cm}(4.75cm,1cm) % {block width} (coords)
%    \centering
%    \textcolor{red}{
%    This work has been submitted to the IEEE International Conference on Robotics and Automation (ICRA) for possible publication. Copyright may be transferred without
% notice, after which this version may no longer be accessible.}
% \end{textblock*}

\maketitle
\thispagestyle{empty}
\pagestyle{empty}

\begin{abstract}
Cinematic camera control demands a balance of precision and artistry—qualities that are difficult to encode through handcrafted reward functions. While reinforcement learning (RL) has been applied to robotic filmmaking, its reliance on bespoke rewards and extensive tuning limits creative usability. We propose a Learning from Demonstration (LfD) approach using Generative Adversarial Imitation Learning (GAIL) to automate dolly-in shots with a free-roaming, ground-based filming robot. Expert trajectories are collected via joystick teleoperation in simulation, capturing smooth, expressive motion without explicit objective design.

Trained exclusively on these demonstrations, our GAIL policy outperforms a PPO baseline in simulation, achieving higher rewards, faster convergence, and lower variance. Crucially, it transfers directly to a real-world robot without fine-tuning—achieving more consistent framing and subject alignment than a prior TD3-based method. These results show that LfD offers a robust, reward-free alternative to RL in cinematic domains, enabling real-time deployment with minimal technical effort. Our pipeline brings intuitive, stylised camera control within reach of creative professionals—bridging the gap between artistic intent and robotic autonomy.
\end{abstract}

\section{Introduction}
Filmmaking demands camera motion that is both precise and expressive. Automating this with ground-based robots introduces challenges that span both technical accuracy and creative sensitivity. When successful, such systems enable consistent, repeatable shots—freeing filmmakers to focus on dynamic and artistically demanding scenes.

Data-driven methods offer promising solutions by sidestepping explicit environmental modeling~\cite{Chen2014}. Among them, reinforcement learning (RL) has shown the ability to learn camera behaviors through trial and error. Recent work has demonstrated automated dolly-in shots with reliable zero-shot sim-to-real transfer on wheeled robots~\cite{Lorimer2024}. Yet despite these advances, ground-based robotic cinematography remains underexplored compared to its aerial counterpart.

However, RL’s practical limitations—handcrafted reward functions, long training times, and high computational cost—present barriers to adoption. Designing aesthetic rewards is difficult, making RL workflows poorly aligned with intuitive, creative filmmaking.

\emph{Learning from Demonstration} (LfD) learns control policies from human camera trajectories, capturing artistic intent without reward engineering. Timing, framing, and composition are encoded in the demos, making LfD effective for cinematography.

While LfD has seen broad application in manipulation and navigation tasks~\cite{Argall2009}, its use in ground-based filmmaking remains limited. Prior work has primarily focused on drones~\cite{Bonatti2020, dang2020enabledronefilmmaker}, while wheeled robots offer unique advantages: stability, precision, and repeatability—key attributes for stylised camera work.

In this paper, we extend the zero-shot sim-to-real framework introduced in~\cite{Lorimer2024}, replacing their RL-based method with an LfD-based pipeline. Our approach removes the need for handcrafted rewards, accelerates training, and better aligns with creative workflows. Policies trained from expert joystick demonstrations in simulation transfer directly to real-world deployment, achieving consistent, expert-like dolly-in shots without fine-tuning.

While our method builds on established imitation learning techniques, our contribution lies in demonstrating that LfD—when applied within this framework—enables robust, accessible cinematographic automation with minimal tuning. Compared to RL, our approach lowers the barrier to entry for both filmmakers and robotics practitioners.

Our key contributions are: \begin{enumerate} \item A complete LfD pipeline for robotic cinematography, from expert data collection to policy training and real-world deployment. \item Quantitative comparison of LfD and RL performance in simulation, relative to expert trajectories. \item Real-world validation demonstrating zero-shot sim-to-real transfer with consistent cinematic behaviour. \end{enumerate}

The remainder of this paper details our problem formulation (Section~\ref{sec:formalisation_of_problem}), data collection and LfD training pipeline (Sections~\ref{sec:data_collection_IL}–\ref{sec:Experimental-pipeline}), experimental setup and results (Section~\ref{sec:Experimental_Results}), and concludes with discussion and future directions (Sections~\ref{sec:discussion_lims}–\ref{sec:conclusions}).

% % ==================
% % PROBLEM FORMULATION
% % ==================
\section{Problem Formulation}
\label{sec:formalisation_of_problem}
The dolly-in shot is a classic cinematographic technique where the camera moves smoothly toward a subject while maintaining precise centring and gradual scaling within the frame. The shot concludes once the subject reaches a desired size. Coordinating motion and framing is a creative task that can be hard to express algorithmically.

% Executing this well requires coordination of robot motion and camera control—an inherently creative task that is difficult to express through rules or objectives.

Reinforcement learning (RL) has been applied to this problem~\cite{Lorimer2024}, but its success depends on carefully engineered reward functions that promote smooth motion, framing stability, and aesthetic composition. These reward signals are often hard to define, especially when the goal is perceptual or stylistic. As a result, RL-based pipelines typically require extensive tuning and domain knowledge, limiting accessibility for filmmakers or creative practitioners.

We propose an alternative: \textit{Learning from Demonstration} (LfD). Rather than explicitly defining success with handcrafted objectives, LfD learns policies directly from expert camera trajectories. These demonstrations naturally encode timing, composition, and stylistic nuance, allowing the robot to reproduce expert-like behaviour without handcrafted rewards.

\begin{figure*}[htbp] \centering \includegraphics[width=0.9\linewidth]{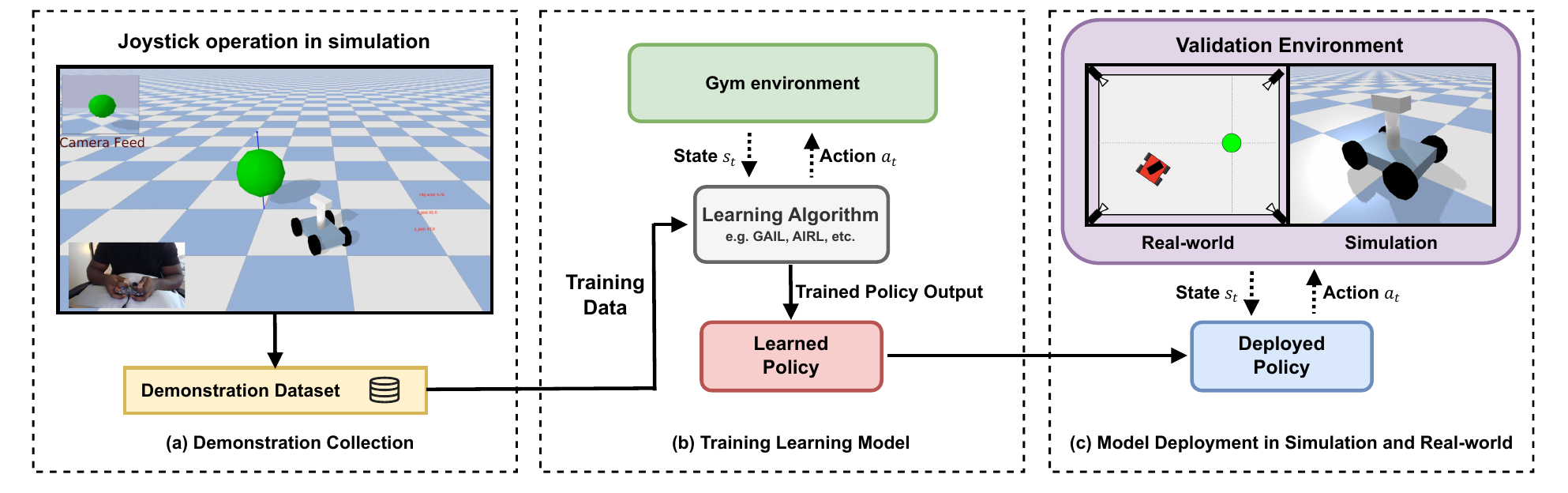} \caption{Overview of the LfD framework comprising three phases: \textit{(a) Demonstration Collection} — expert trajectories are recorded via joystick teleoperation in simulation; \textit{(b) Training} — An LfD algorithm learns a policy from these demonstrations; \textit{(c) Deployment} — the policy is deployed in simulation and the real world for autonomous camera control.} \label{fig:Lfd_pipeline} \end{figure*}

Our method follows a three-stage Learning from Demonstration pipeline: expert demonstration, policy training, and deployment. This is illustrated in Figure~\ref{fig:Lfd_pipeline}, which outlines the full flow from joystick-operated data collection to simulation and real-world evaluation. To enable direct comparison with prior work, we adopt the same simulation environment, robot models, and camera setup as in~\cite{Lorimer2024}, but replace the RL core with an LfD-based approach.

% ======================================
% ======================================

% Demonstration collection
\section{Demonstration Collection}
\label{sec:data_collection_IL}

\begin{figure}[htbp] \centering \includegraphics[width=0.9\linewidth]{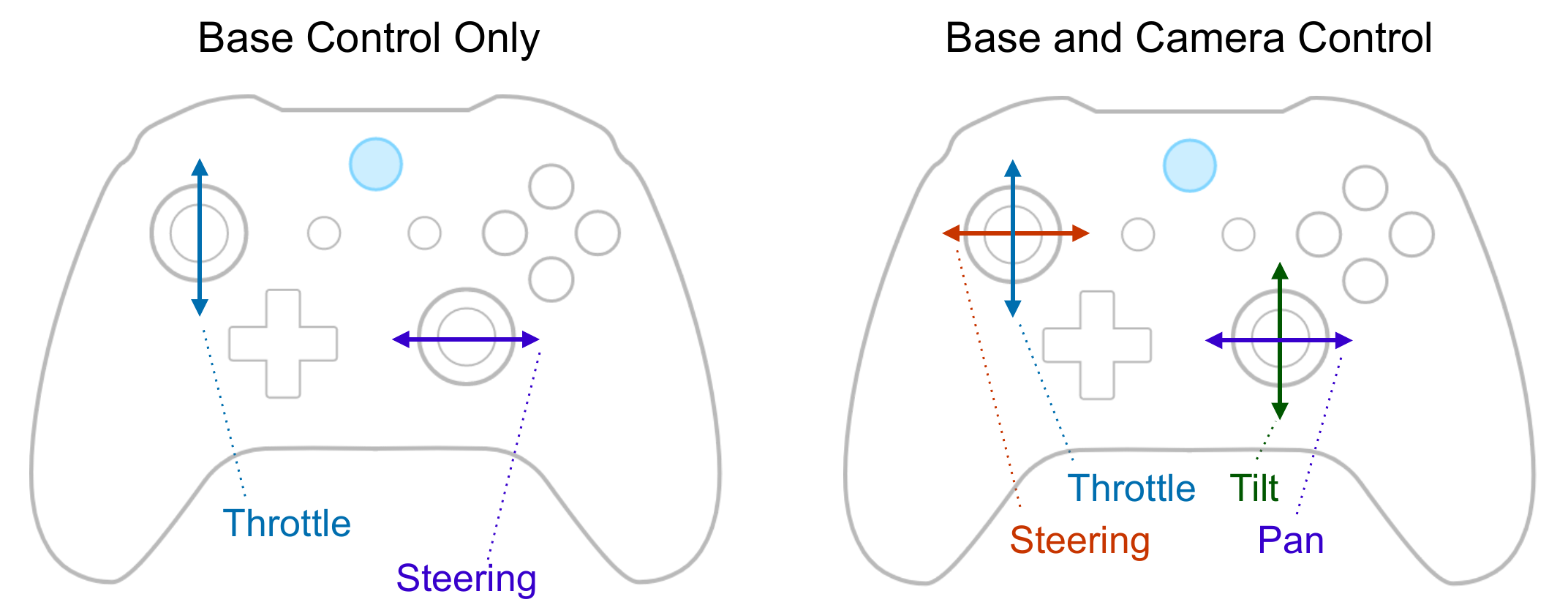} \caption{Joystick control interface setup, showing the Xbox controller integrated with a laptop running the cinematography simulation environment.} \label{fig:control_interface} \end{figure}

To train our Learning from Demonstration (LfD) pipeline, we collected expert camera trajectories in simulation via joystick teleoperation. A single operator controlled a ground robot with a virtual camera using an Xbox controller, interfaced through Pygame~\cite{pygame} in a PyBullet-based simulation environment~\cite{coumans2019, brockman2016openai}. This setup enabled expressive, real-time control with minimal training overhead, and reflects standard practice in LfD, where teleoperation is commonly used to capture expert behaviour~\cite{Argall2009, osa2018algorithmic} (Figure~\ref{fig:control_interface}).

Each demonstration recorded a complete dolly-in trajectory as a sequence of state-action pairs. We collected 25 demonstrations per task (Base and Full Control), varying the robot’s starting position, orientation, and lighting to promote generalisation. Data were logged using the \texttt{TrajectoryAccumulator} from the Imitation Library~\cite{gleave2022imitation}, producing standardised datasets of observations, actions, and transitions.

\begin{figure}[htbp] \centering \includegraphics[width=0.9\linewidth]{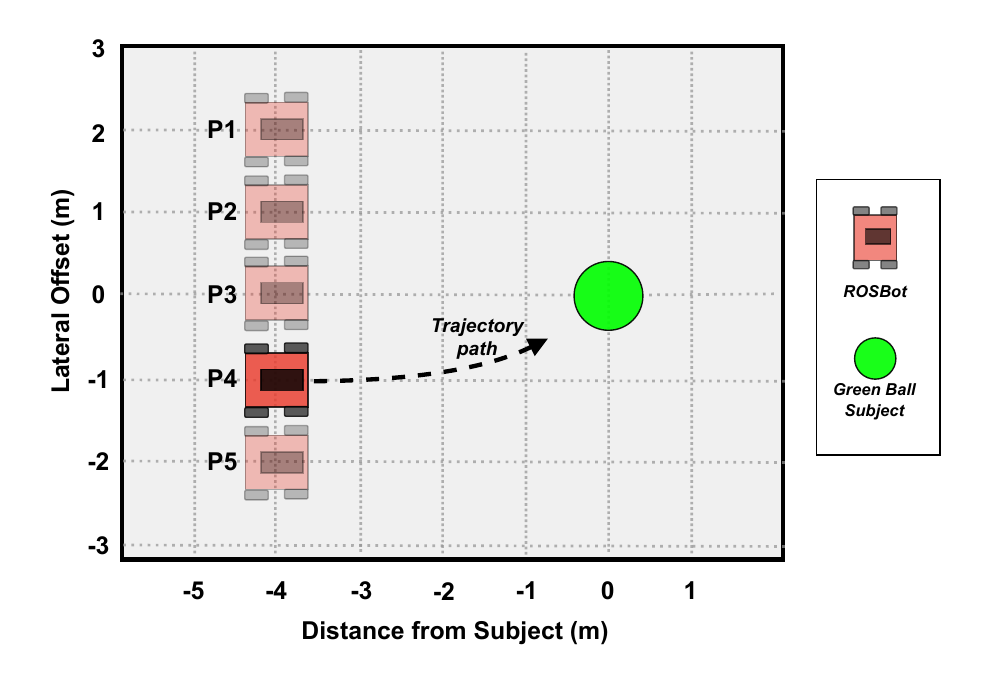} \caption{ Robot starting positions for demonstration diversity. Positions P1–P5 span left to right across the scene. Diversity levels were defined by the number of positions used: low (P3 only), moderate (P1, P3, P5), and high (P1–P5). This setup enabled controlled evaluation of how spatial variation affects generalisation.} \label{fig:robot_starting_positions} \end{figure}

\begin{figure}[htbp] \centering \includegraphics[width=0.9\linewidth]{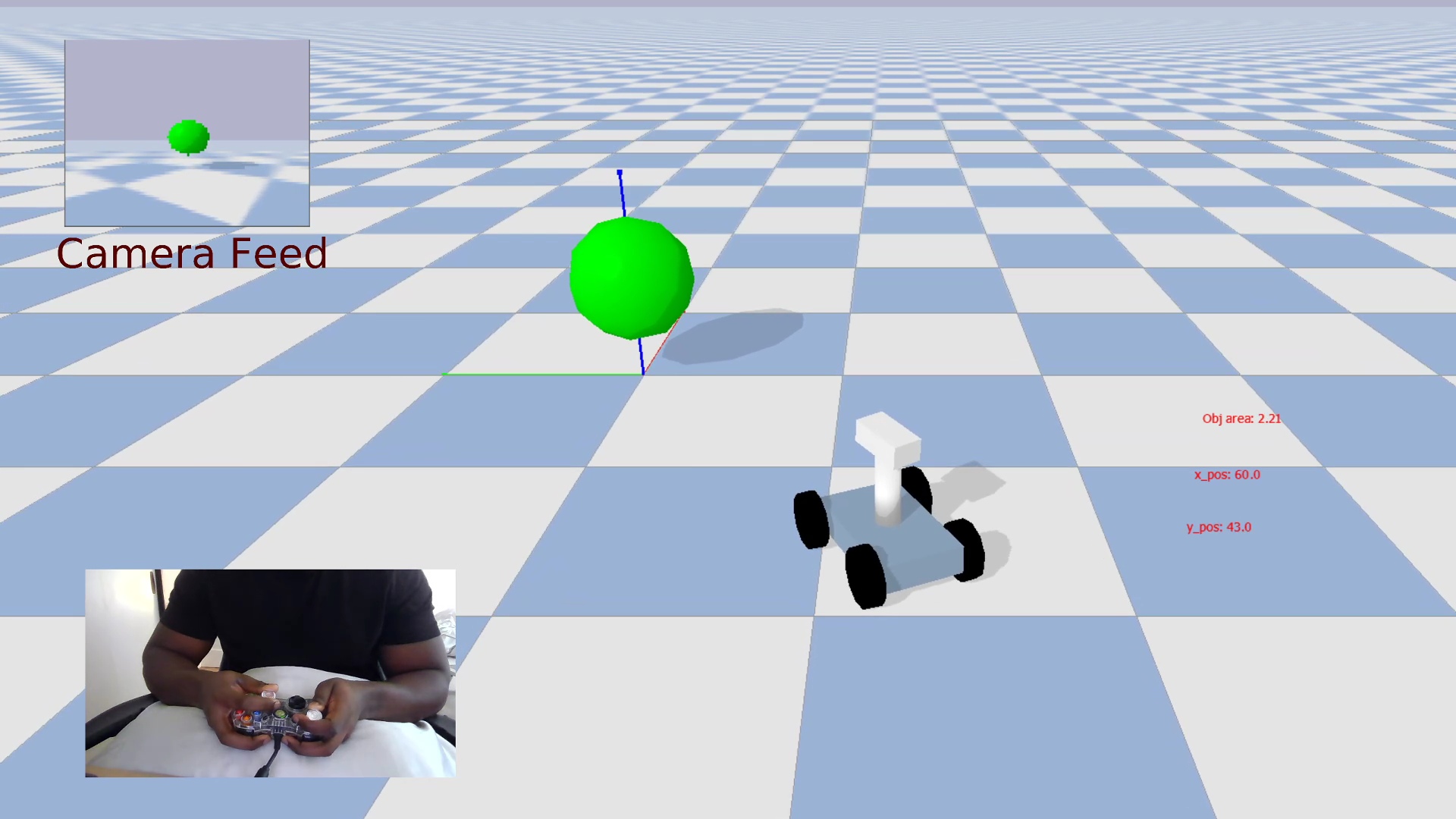} \caption{Example view of the PyBullet-based simulation environment used to collect expert cinematographic demonstrations.} \label{fig:simulation_environment} \end{figure}

To ensure consistency, a single expert operator performed all demonstrations. The spatial distribution of initial positions is shown in Figure~\ref{fig:robot_starting_positions}, with a representative view of the simulation environment in Figure~\ref{fig:simulation_environment}.

 \section{Learning from Demonstration (LfD)}
\label{sec:Experimental-pipeline} 
Robotic cinematography presents a challenge: artistic intent is difficult to encode explicitly. While reinforcement learning (RL) has been applied to automate camera motion, it depends on handcrafted rewards that are often subjective and costly to design. In contrast, \textit{Learning from Demonstration (LfD)} enables policies to learn directly from expert trajectories—capturing framing, timing, and style without manual reward design.

We model the task as a Markov Decision Process (MDP) with states \(S\), actions \(A\), transitions \(T\), and discount factor \(\gamma\), where the reward \(R\) is unknown or hard to define. Instead, we assume access to expert trajectories \(\tau = \{(s_0, a_0), \dots, (s_n, a_n)\}\), from which the goal is to recover a policy \(\pi(a|s)\) that mimics expert behaviour.

We evaluate two learning strategies: Proximal Policy Optimisation (PPO), a reinforcement learning baseline trained with handcrafted rewards, and Generative Adversarial Imitation Learning (GAIL), our primary LfD method, trained solely on expert demonstrations without access to rewards.

\subsection{PPO Baseline}
\textit{Proximal Policy Optimisation (PPO)}~\cite{schulman2017proximalpolicyoptimizationalgorithms} is an on-policy RL algorithm optimising the clipped objective:
\[
L^{\text{CLIP}}(\theta) = \hat{\mathbb{E}}_t\left[\min\left(r_t(\theta)\hat{A}_t, \text{clip}(r_t(\theta), 1 \! - \! \epsilon, 1 \! + \! \epsilon)\hat{A}_t\right)\right],
\]
where \(r_t(\theta)\) is the policy ratio and \(\hat{A}_t\) the advantage. We use the Stable Baselines 3 implementation~\cite{stable-baselines3} with the same reward structure as in.~\cite{Lorimer2024} (see Section~\ref{sec:Experimental-Setup}). While TD3 performed well in previous work, we use PPO as our RL baseline due to its simplicity, widespread use, and compatibility with the on-policy imitation framework provided by the \texttt{imitation} library~\cite{gleave2022imitation}.

% Although TD3~\cite{Lorimer2024} performed well, it required extensive tuning and reward shaping. PPO, by contrast, is simpler, widely adopted, and compatible with our on-policy imitation setup. It thus serves as a reproducible and consistent RL benchmark.

\subsection{GAIL for Cinematic Imitation}
\textit{Generative Adversarial Imitation Learning (GAIL)}~\cite{GAIL2016} casts imitation as a two-player game:
\begin{enumerate}
    \item A \textbf{discriminator} \(D_\phi(s, a)\) distinguishes expert from agent behaviour.
    \item A \textbf{policy} \(\pi_\theta(a|s)\) learns to fool the discriminator.
\end{enumerate}
The training objective is:
\[
\min_{\pi_\theta} \max_{D_\phi} \mathbb{E}_{\pi_E}[\log D_\phi(s, a)] + \mathbb{E}_{\pi_\theta}[\log(1 - D_\phi(s, a))].
\]

We implement GAIL in PyTorch using the \texttt{imitation} library~\cite{gleave2022imitation}, with learning rate \(1 \times 10^{-4}\) and batch size 64. The learned reward signal allows the agent to align with expert behaviour without explicit design.

Although we evaluated newer methods (e.g., LS-IQ, IQ-Learn), they proved unstable under our constraints. GAIL offered greater robustness and convergence, making it the most suitable option.

Prior work has validated GAIL in manipulation and construction~\cite{GAIL_in_construction, GAIL_manipulation}; we extend it to camera control, which requires smooth, perceptual, and temporally consistent outputs—properties well suited to GAIL’s formulation.

% ===================================
% ===================================

\section{Experimental Results and Analysis}
\label{sec:Experimental_Results}
This section details the experimental setup and outcomes, validating the effectiveness of RL and LfD agents in performing autonomous dolly-in shot tasks.

\subsection{Experimental Setup}
\label{sec:Experimental-Setup}

We evaluate our Learning from Demonstration (LfD) pipeline in the high-fidelity PyBullet simulation environment introduced in~\cite{Lorimer2024}, which simulates a ground-based filming robot with a controllable camera. To ensure direct comparability, we adopt the same robot model, environment, and reward structure used in PPO training. Our contribution is the integration of an LfD pipeline based on expert joystick demonstrations.

\paragraph*{Task Description}
The target behaviour is a dolly-in shot: the robot moves toward a static subject while keeping it centered and smoothly scaled within the frame. This requires coordinated locomotion and camera control.

We evaluate two task variants:
\begin{enumerate}
    \item \textbf{Base Control}: Controls throttle and steering.
    \item \textbf{Full Control}: Adds pan and tilt for active framing.
\end{enumerate}

\paragraph*{Action Space}
Depending on the control mode, the policy outputs either:
\[
a = 
\begin{cases}
[\text{throttle}, \text{steering}] & \text{(Base)} \\
[\text{throttle}, \text{steering}, \text{pan}, \text{tilt}] & \text{(Full)}
\end{cases}
\]
All actions are scaled to respect physical limits and passed through \texttt{tanh} activations for smooth transitions.

\paragraph*{PPO Reward Function}
The PPO baseline is trained using a scalar reward adapted from ~\cite{Lorimer2024}, combining two terms:

\begin{itemize}
    \item \textbf{Framing Progress}: Rewards forward motion that increases the subject's size in the frame while keeping it centered.
    \item \textbf{Motion Smoothness}: Penalises abrupt changes in control inputs to promote cinematic stability.
\end{itemize}

For the \textbf{Base Control} task (throttle and steering), the reward at each timestep is:
\[
r_t = \lambda_{\text{area}} \cdot \Delta A_t - \lambda_{\text{steer}} \cdot \Delta \dot{\theta}_t^2
\]
where \( \Delta A_t \) is the change in object area (subject scale) and \( \Delta \dot{\theta}_t \) is the change in steering rate.

For the \textbf{Full Control} task (adds pan and tilt), the smoothness term is extended to include camera motion:
\[
r_t = \lambda_{\text{area}} \cdot \Delta A_t - \lambda_{\text{steer}} \cdot \Delta \dot{\theta}_t^2 - \lambda_{\text{cam}} \cdot (\Delta \dot{\phi}_t^2 + \Delta \dot{\psi}_t^2)
\]
where \( \Delta \dot{\phi}_t \) and \( \Delta \dot{\psi}_t \) are changes in pan and tilt rates, respectively.

Hyperparameters \( \lambda_{\text{area}}, \lambda_{\text{steer}}, \lambda_{\text{cam}} \) are tuned via grid search to balance framing accuracy and motion stability.

\paragraph*{Training Protocol}
Each agent is trained for 1 million timesteps using 1500-step episodes. We run three random seeds per configuration to assess variance. Evaluation is based on final episodic reward (100 trials), convergence speed, and stability.

\subsection{Simulation Experiments}
We compare Generative Adversarial Imitation Learning (GAIL) and Proximal Policy Optimisation (PPO) on the dolly-in cinematography task across two settings: \textbf{Base Control} (throttle and steering) and \textbf{Full Control} (adds pan and tilt). GAIL is trained on 25 joystick-operated demonstrations, while PPO learns from scratch using the handcrafted reward defined in~\cite{Lorimer2024}. For additional context, we include the TD3 results from the same work as a high-performing, reward-engineered baseline requiring substantially more training.
% We evaluate the effectiveness of Generative Adversarial Imitation Learning (GAIL) compared to Proximal Policy Optimization (PPO) on the dolly-in cinematography task, across two settings: \textbf{Base Control} (throttle and steering) and \textbf{Full Control} (adds pan and tilt). GAIL is trained using 25 joystick-operated expert demonstrations, while PPO learns from scratch using the handcrafted reward from Lorimer et al.~\cite{Lorimer2024}. For reference, we include the TD3 agent from Lorimer et al.~\cite{Lorimer2024}, which performs well but requires significantly more training.

\paragraph{Overall Performance.} As shown in Table~\ref{tab:gail_vs_ppo_rewards}, GAIL consistently outperforms PPO across both task settings. In Base Control, GAIL achieves an average reward improvement of \textbf{8.4\%}, and in Full Control, a \textbf{4.3\%} gain. GAIL also exhibits faster convergence and lower variance across three training seeds. Although TD3 achieves the highest reward, it is considerably less sample efficient.

\begin{figure}[ht]
\centering
\includegraphics[width=\linewidth]{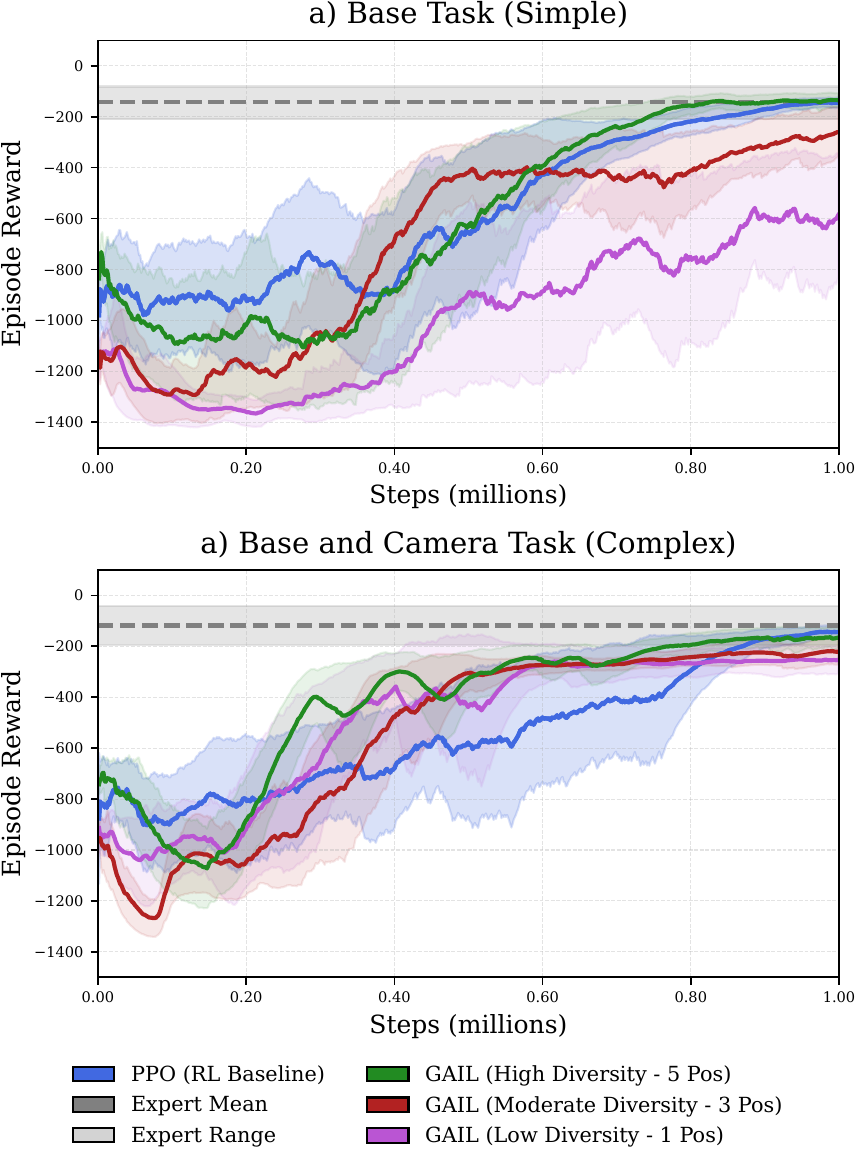}
\caption{
 Learning curves comparing PPO (RL baseline) and GAIL (LfD) on (a) the Base Task and (b) the Full Task. GAIL is trained on demonstrations with increasing diversity (1, 3, or 5 start positions). Curves show mean episodic reward for 3 seeds over 1m training steps; shaded regions represent ±1 standard deviation. The dashed line and band show the expert’s mean and range. Results highlight that increased demonstration diversity improves GAIL’s consistency and overall performance, enabling it to match or exceed PPO}
\label{fig:gail_vs_ppo_diversity_learning}
\end{figure}

\paragraph{Impact of Demonstration Diversity.} To assess how the diversity of demonstrations affects policy learning, we compare GAIL agents trained on 25 demonstrations sampled from \textit{low} (1), \textit{moderate} (3), and \textit{high} (5) distinct starting positions. Figure~\ref{fig:gail_vs_ppo_diversity_learning} presents the resulting learning curves. Greater start position diversity leads to improved generalisation and more stable learning. In both tasks, high-diversity GAIL matches or outperforms PPO, underscoring that not only the \textit{quantity} but the \textit{distribution} of demonstrations is crucial for robust behavior.

\begin{table}[ht]
\centering
\caption{
Comparison of GAIL and PPO on dolly-in shots. GAIL outperforms PPO in both task settings using only 25 expert demonstrations. TD3 performs best but requires significantly more experience, highlighting the trade-off between sample efficiency and performance.
}
\begin{tabular}{lcc}
\toprule
\textbf{Method} & \textbf{Base Control} & \textbf{Full Control} \\
\midrule
\textbf{GAIL (25 Demos)} & \textbf{-116.3 ± 36.0} & \textbf{-126.3 ± 25.6} \\
PPO Baseline             & -127 ± 25.0              & -132 ± 24.1 \\
% TD3 (Lorimer et al.)     & -107.8 ± 24.1          & -107.8 ± 24.1 \\
\midrule
Expert Demonstrations    & -142.1 ± 29.5          & -118.1 ± 39.8 \\
\bottomrule
\end{tabular}

\label{tab:gail_vs_ppo_rewards}
\end{table}

% \subsubsection{Experiment 2: Effect of Demonstration Diversity}

% We evaluate how the diversity of expert demonstrations influences policy performance. All GAIL agents are trained with 25 demonstrations, but drawn from varying numbers of starting configurations:
% \begin{itemize}
%     \item \textbf{Low diversity}: 1 start position
%     \item \textbf{Moderate diversity}: 3 start positions
%     \item \textbf{High diversity}: 5 start positions
% \end{itemize}

% As shown in Figure~\ref{fig:diversity_learning_curves}, higher diversity significantly improves convergence speed and final reward, especially under Full Control. Only the high-diversity setting reliably achieved expert-level performance, highlighting the importance of varied data for robust generalisation.

% These results underscore that not just the \textit{quantity}, but the \textit{diversity} of demonstrations is crucial for learning generalizable cinematic behaviour. Full training curves and additional metrics are provided in the Supplementary Material.

% \begin{figure}[htbp]
%     \centering
%     \includegraphics[width=\linewidth]{images/temp_results_diversity_updated.pdf}
%     \caption{
%     Learning curves for GAIL under different demonstration diversity conditions. Policies trained with \textbf{high diversity} (five start positions) show faster, more stable convergence to expert-level performance.
%     }
%     \label{fig:diversity_learning_curves}
% \end{figure}

% \subsubsection{Experiment 3: Effect of Demonstration Quantity}

\subsection{Real-world Experiments}
To validate zero-shot Sim2Real transfer, we deploy the GAIL (PPO) policy, trained entirely in simulation using 25 diverse demonstrations, onto a physical ground robot with no fine-tuning. This tests the system’s ability to generalise cinematic behaviour under real-world conditions, mirroring the simulation setup in Experiment 1.

Following ~\cite{Lorimer2024}, we evaluate each method from three canonical starting positions (left, centre, right), recording cumulative reward and camera framing metrics (object area and X/Y position). As in simulation, rewards are computed per step, and framing targets remain aligned with expert intent. Sim2Real fidelity is quantified using the Sim2Real Correlation Coefficient (SRCC, \cite{kadian2020}), measuring alignment between simulated and real-world outcomes.

Results (Table~\ref{tab:real2sim_experiment_comparison_chpt2}) show that GAIL (PPO) consistently outperforms the TD3 baseline across all start positions. GAIL achieves higher cumulative rewards and significantly stronger SRCC scores, including near-perfect correlation on object area ($\geq$0.97). Additional error breakdowns (Tables~\ref{tab:area_errors}–\ref{tab:pos_y_error}) show improvements of up to \textbf{100\%} in object area accuracy and over \textbf{88\%} in horizontal centering.

To test generalisation beyond seen configurations, we conduct 75 additional trials from randomised positions. The robot successfully completes the dolly-in in all 75 trials (100\%) where the subject is visible, demonstrating robust generalisation and cinematic framing under variation.

\begin{figure}[h]
  \centering

  \begin{subfigure}[b]{\linewidth}
    \centering
    \includegraphics[width=\linewidth]{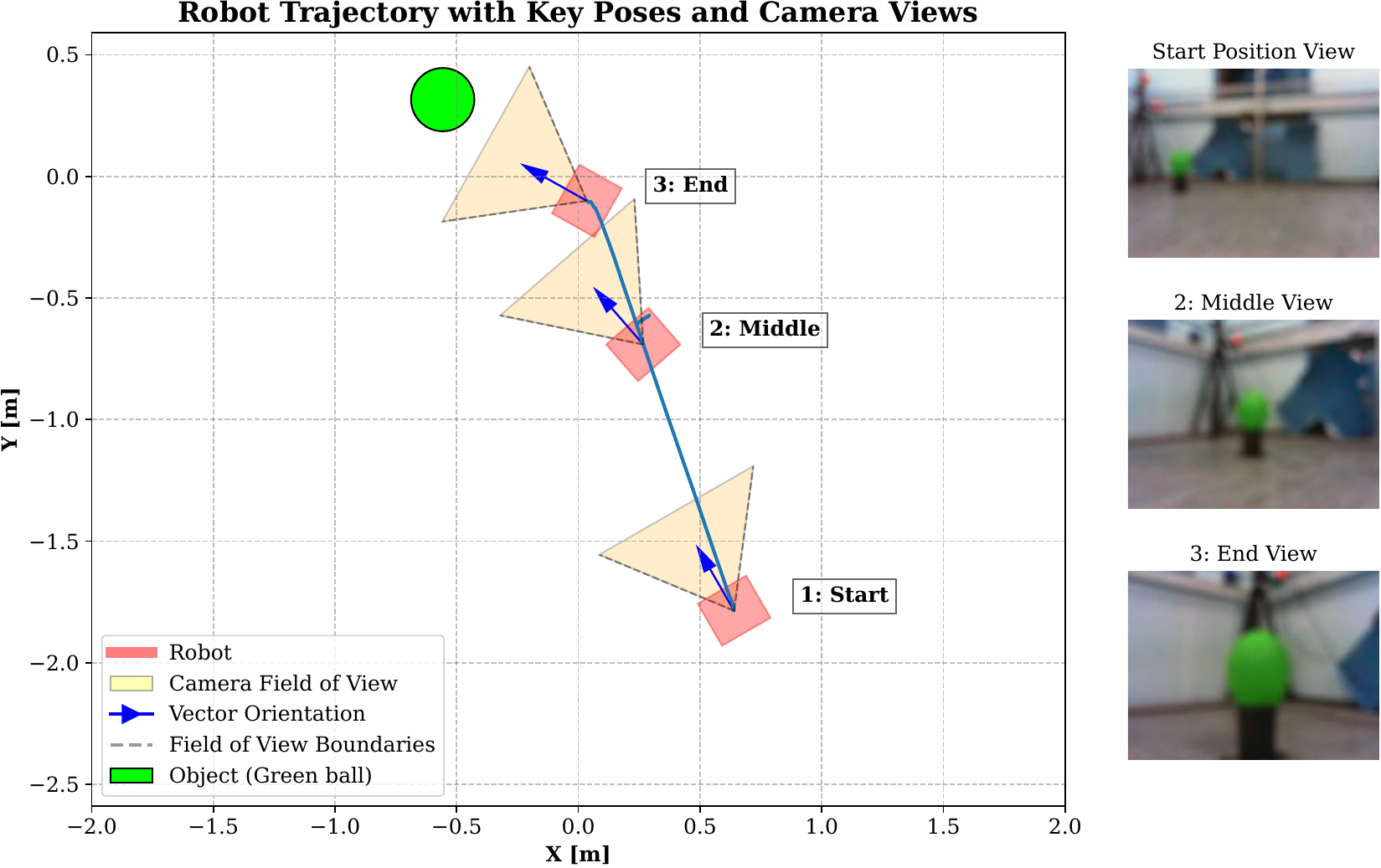}
    \caption{Single execution from one start. Heading vectors show smooth control and subject framing.}
    \label{fig:single-trajectory-visual}
  \end{subfigure}

  \vspace{0.5cm}

  \begin{subfigure}[b]{\linewidth}
    \centering
    \includegraphics[width=0.8\linewidth]{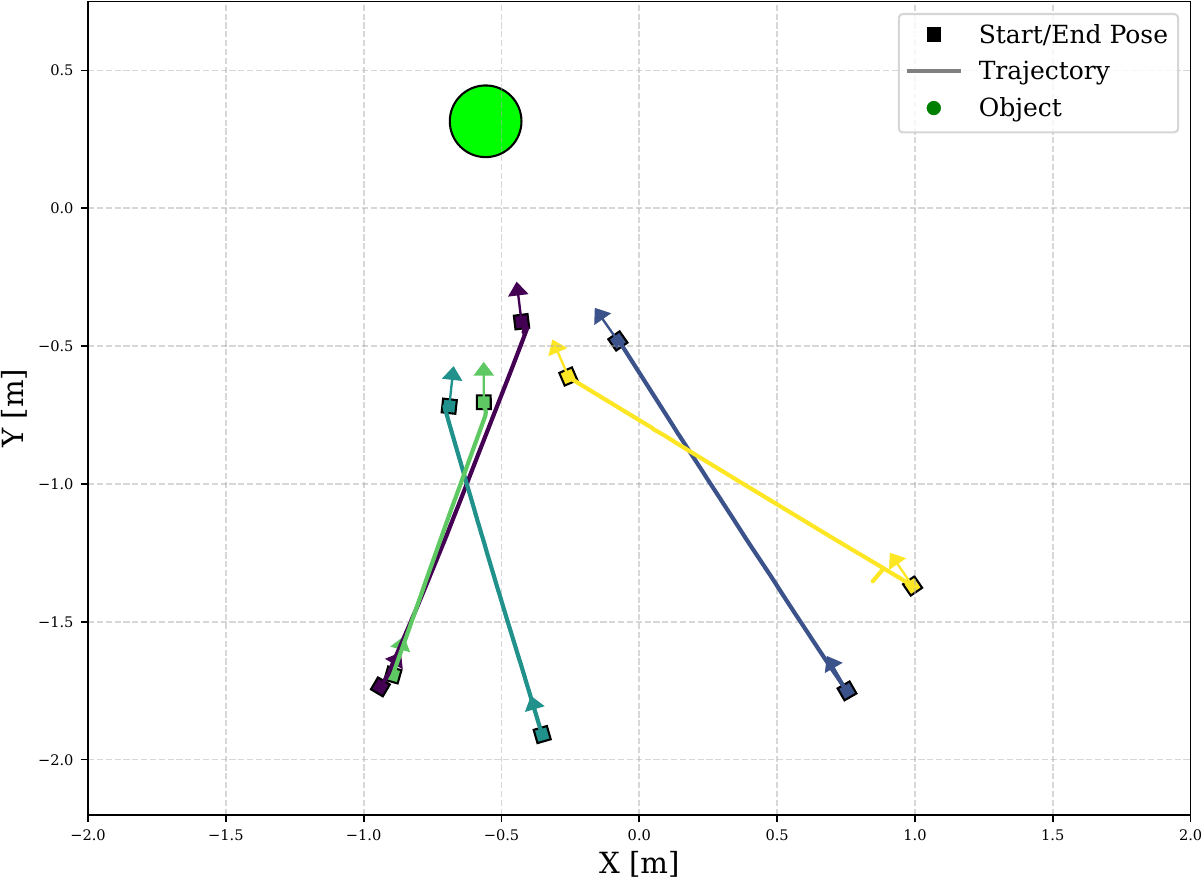}
    \caption{Overlaid runs from a different start. The robot exhibits consistent behaviour across trials.}
    \label{fig:multi-trajectory-visual}
  \end{subfigure}

  \caption{Real-world dolly-in trajectory visualisations. (a) shows a single execution with heading vectors overlaid, highlighting smooth subject framing and motion. (b) overlays multiple trajectories from a different start position, demonstrating consistent, repeatable behaviour. These results confirm that the GAIL (PPO) policy generalises well to physical deployments with high cinematic quality.}
  \label{fig:trajectory-combined}
\end{figure}

Figure~\ref{fig:trajectory-combined} visualises typical real-world trajectories. Subfigure (a) shows one trajectory with heading vectors overlaid, while subfigure (b) overlays several runs from a separate start location, illustrating consistency across executions. Camera views confirm that the subject remains centred and appropriately scaled, validating cinematic quality.

These results validate that cinematic policies learned via imitation in simulation can reliably generalise to real-world deployment, reducing the need for domain-specific tuning or reward design.

\begin{figure}[h]
  \centering
  \includegraphics[width=0.85\linewidth]{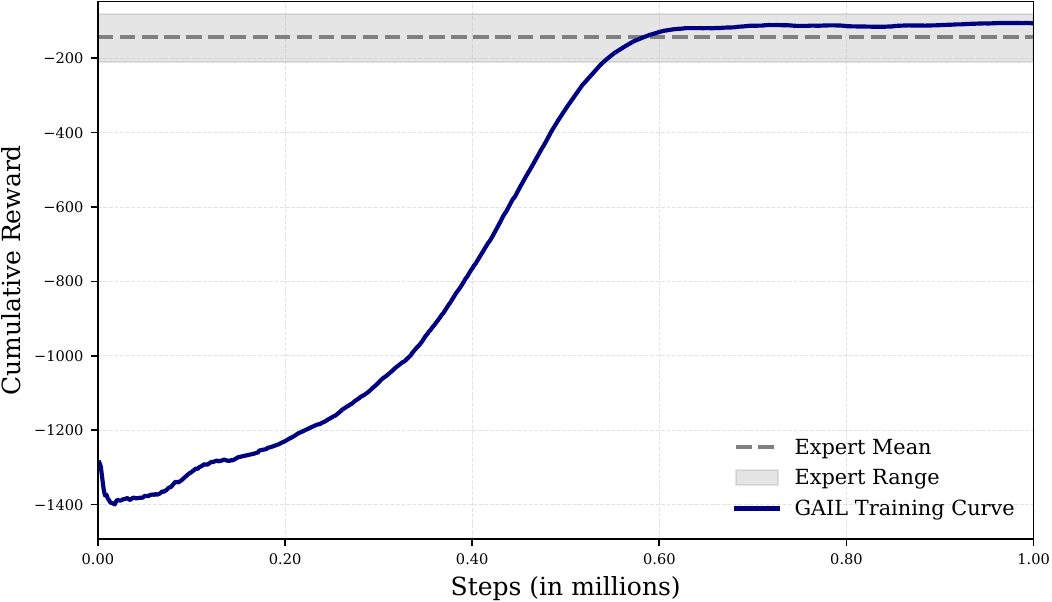}
  \caption{Simulation training curve for the GAIL (PPO) policy used in real-world deployment. The policy converges to a reward level consistent with expert demonstrations, supporting its readiness for zero-shot transfer without fine-tuning.}
  \label{fig:sim2real_training}
\end{figure}

% =================================
% =================================

% Define gradient color ranges
\definecolor{lowcolor}{HTML}{FFCCCC} % Light red for poor/weak correlations
\definecolor{low-midcolor}{HTML}{FFFFCC} % Light yellow for moderate correlations
\definecolor{midcolor}{HTML}{DDFFDD} % Light green for strong correlations
\definecolor{highcolor}{HTML}{99FF99} % Bright green for very strong correlations
\definecolor{negcolor}{HTML}{FF9999} % Darker red for negative correlations

\begin{table*}[hpbt!]
\centering

\caption{
Sim2Real performance comparison between the \textbf{TD3 baseline~\cite{Lorimer2024}} and our proposed \textbf{GAIL (PPO)} method, across start positions. Metrics include cumulative reward, object area, and X/Y position in both simulation and real-world deployments. The Sim2Real Rank Correlation Coefficient (SRCC) quantifies consistency between simulated and real-world outcomes. \textit{SRCC values are colour-coded as follows: bright green (very strong: 0.8–1), light green (strong: 0.6–0.79), light yellow (moderate: 0.4–0.59), light red (weak: 0–0.39), and darker red (negative: $\leq$0)}. Higher SRCC indicates better transferability. This table summarises key performance outcomes supported by detailed error breakdowns in Tables~\ref{tab:area_errors}–\ref{tab:pos_y_error}.
}
\resizebox{\textwidth}{!}{%
\begin{tabular}{ll*{2}{c} *{3}{c} *{3}{c} *{3}{c}}
\toprule
& & \multicolumn{2}{c}{Cum. Reward} & \multicolumn{3}{c}{Object Area} & \multicolumn{3}{c}{Object Position (X-axis)} & \multicolumn{3}{c}{Object Position (Y-axis)} \\
\cmidrule(lr){3-4} \cmidrule(lr){5-7} \cmidrule(lr){8-10} \cmidrule(lr){11-13}
Method & Start Pos. & Sim. & Real. & Sim. & Real. & SRCC & Sim. & Real & SRCC & Sim. & Real & SRCC \\
\midrule
TD3 Baseline & Left   & -170.77 & -176.27 & 10.12 & 9.63 & \cellcolor{midcolor}0.69 & 59.42 & 61.22 & \cellcolor{low-midcolor}0.52 & 44.0 & 44.54 & \cellcolor{midcolor}\textbf{0.65} \\
                                          & Right  & -166.03 & -166.35 & 10.09 & 10.38 & \cellcolor{highcolor}0.80 & 59.32 & 59.08 & \cellcolor{midcolor}0.72 & 44.1 & 44.23 & \cellcolor{highcolor}\textbf{0.83} \\
                                          & Centre & -129.93 & -134.96 & 10.09 & 11.09 & \cellcolor{low-midcolor}0.46 & 59.79 & 55.67 & \cellcolor{low-midcolor}0.56 & 44.0 & 46.51 & \cellcolor{midcolor}\textbf{0.69} \\
                                    \midrule
GAIL (PPO)                        & Left   & \textbf{-145.19} & \textbf{-143.07} & 10.23 & 10.20 & \cellcolor{highcolor}\textbf{0.97} & 59.82 & 59.59 & \cellcolor{midcolor}\textbf{0.72} & 38.92 & 37.22 & \cellcolor{low-midcolor}0.58 \\
                                          & Right  & \textbf{-130.95} & \textbf{-132.16} & 10.20 & 10.0 & \cellcolor{highcolor}\textbf{0.98} & 59.78 & 59.88 & \cellcolor{midcolor}\textbf{0.76} & 38.96 & 37.26 & \cellcolor{midcolor}0.61 \\
                                          & Centre & \textbf{-117.60} & \textbf{-116.05} & 10.16 & 9.84 & \cellcolor{highcolor}\textbf{0.99} & 59.76 & 59.51 & \cellcolor{highcolor}\textbf{0.87} & 38.96 & 36.46 & \cellcolor{midcolor}0.67 \\
\bottomrule
\end{tabular}%
}

\label{tab:real2sim_experiment_comparison_chpt2}
\end{table*}

\definecolor{lowcolor}{HTML}{FFCCCC}   % High Error (red)
\definecolor{midcolor}{HTML}{FFFFCC}   % Moderate Error (yellow)
\definecolor{highcolor}{HTML}{99FF99}  % Low Error (green)

\begin{table}[pbt!]
\centering
\renewcommand{\arraystretch}{1.3}
\caption{
Object area error across start positions. GAIL (PPO) achieves lower error in all cases, including a perfect match at the right position. Values show absolute deviation from the 10-unit framing target, with percentages indicating relative error. Colours indicate severity: green (low), yellow (moderate), red (high).
}
\resizebox{\linewidth}{!}{
\begin{tabular}{l ccc}
\toprule
\multirow{2}{*}{Start Pos.} 
& \multicolumn{3}{c}{Object Area Error} \\
\cmidrule(lr){2-4}
& TD3 & GAIL (PPO) & \% Improv. \\ 
\midrule
Left
& \cellcolor{midcolor}0.37 (\textit{3.7\%})  
& \cellcolor{highcolor}\textbf{0.20} (\textit{2.0\%})  
& +45.9\%  \\

Right
& \cellcolor{midcolor}0.38 (\textit{3.8\%})  
& \cellcolor{highcolor}\textbf{0.00} (\textit{0.0\%})  
& +100.0\%  \\

Centre
& \cellcolor{lowcolor}1.09 (\textit{10.9\%})  
& \cellcolor{midcolor}\textbf{0.16} (\textit{1.6\%})  
& +85.3\%  \\
\bottomrule
\end{tabular}
}

\label{tab:area_errors}
\end{table}

\definecolor{lowcolor}{HTML}{FFCCCC}   % High Error (red)
\definecolor{midcolor}{HTML}{FFFFCC}   % Moderate Error (yellow)
\definecolor{highcolor}{HTML}{99FF99}  % Low Error (green)

\begin{table}[pbt!]
\centering
\renewcommand{\arraystretch}{1.3}

\caption{
Horizontal (X-axis) alignment error across start positions. GAIL (PPO) outperforms the TD3 baseline, reducing error by over 88\% in the most challenging scenario. Errors are shown relative to the target screen centre (60).
}

\resizebox{\linewidth}{!}{
\begin{tabular}{l ccc}
\toprule
\multirow{2}{*}{Start Pos.}
& \multicolumn{3}{c}{X Position Error} \\
\cmidrule(lr){2-4}
& TD3 & GAIL (PPO) & \% Improv. \\
\midrule
Left  
& \cellcolor{lowcolor}1.22 (\textit{2.03\%})  
& \cellcolor{midcolor}\textbf{0.41} (\textit{0.68\%})  
& +66.4\%  \\

Right
& \cellcolor{midcolor}0.92 (\textit{1.53\%})  
& \cellcolor{highcolor}\textbf{0.12} (\textit{0.20\%})  
& +87.0\%  \\

Centre
& \cellcolor{lowcolor}4.33 (\textit{7.22\%})  
& \cellcolor{midcolor}\textbf{0.49} (\textit{0.82\%})  
& +88.7\%  \\
\bottomrule
\end{tabular}
}

\label{tab:pos_x_error}
\end{table}

\definecolor{lowcolor}{HTML}{FFCCCC}   % High Error (red)
\definecolor{midcolor}{HTML}{FFFFCC}   % Moderate Error (yellow)
\definecolor{highcolor}{HTML}{99FF99}  % Low Error (green)

\begin{table}[pbt!]
\centering
\renewcommand{\arraystretch}{1.3}

\caption{
Vertical (Y-axis) alignment error from the target position (40) across start positions. GAIL (PPO) substantially improves vertical centring compared to TD3, especially from the centre start. Error values reflect absolute deviation, with colour highlighting the severity.
}

\resizebox{\linewidth}{!}{
\begin{tabular}{l ccc}
\toprule
\multirow{2}{*}{Start Pos.}
& \multicolumn{3}{c}{Y Position Error} \\
\cmidrule(lr){2-4}
& TD3 & GAIL (PPO) & \% Improv. \\
\midrule
\textbf{Left}   
& \cellcolor{lowcolor}4.54 (\textit{11.35\%}) 
& \cellcolor{midcolor}\textbf{2.78} (\textit{6.95\%})  
& +38.8\%  \\

Right
& \cellcolor{lowcolor}4.23 (\textit{10.58\%}) 
& \cellcolor{midcolor}\textbf{2.74} (\textit{6.85\%})  
& +35.2\%  \\

Centre
& \cellcolor{lowcolor}6.51 (\textit{16.28\%}) 
& \cellcolor{midcolor}\textbf{3.54} (\textit{8.85\%})  
& +45.6\%  \\
\bottomrule
\end{tabular}
}

\label{tab:pos_y_error}
\end{table}

\section{Discussion and Limitations}
\label{sec:discussion_lims}

Our results show that Learning from Demonstration (LfD), implemented via GAIL, offers a practical alternative to reinforcement learning (RL) for robotic cinematography. GAIL-trained policies, using only 25 demonstrations, outperform PPO in simulation and transfer directly to real hardware without fine-tuning—removing the need for reward design or extensive tuning. This lowers the barrier for creative users and enables rapid deployment of stylised robotic behaviour.

% \subsection*{Simulation Assumptions}
% In simulation, GAIL shows faster convergence, higher rewards, and lower variance than PPO, especially with diverse demonstrations. However, data was collected from a single expert, which may limit stylistic variability. The simulator also abstracts away noise, occlusion, and dynamic scenes, which future work could address using richer environments or style-conditioned policies.

In simulation, GAIL outperforms PPO, faster convergence, higher rewards, lower variance, especially with diverse demos. Caveats: single-expert data and a simplified simulator lacking noise, occlusions, and dynamics; next steps include richer environments or style-conditioned policies

% \subsection*{Real-world Constraints}
Real-world tests confirm zero-shot transfer under controlled conditions: flat ground, static subject, and consistent lighting. Performance in more dynamic scenarios (e.g., subject re-identification, occlusions) remains untested. Despite low-cost sensors and actuators, GAIL handled real-world noise well—suggesting robustness—but higher-end hardware could unlock more advanced behaviours like adaptive framing or camera motion.

% \subsection*{Methodological Trade-offs}
While GAIL avoids reward tuning, it introduces adversarial training challenges. Compared to behavioural cloning (BC), it requires more hyperparameter tuning and stable demonstrations. For longer or more complex tasks, data collection may become a bottleneck. Tooling for streamlined demonstration capture or hybrid LfD–RL pipelines could improve scalability and creative flexibility.

\section{Conclusions and Future work}
\label{sec:conclusions}

e presented a data-driven pipeline for robotic cinematography using Learning from Demonstration (LfD), enabling ground-based robots to perform expert-level dolly-in shots without handcrafted rewards. Using joystick-operated demonstrations, we trained GAIL policies in simulation that transferred successfully to real-world deployment in a zero-shot manner—achieving consistent, cinematic behavior with no fine-tuning.

Compared to reinforcement learning, our approach improves learning efficiency, reduces engineering effort, and aligns better with creative workflows. Real-world results confirm strong performance and high simulation-to-deployment fidelity, supporting its practical utility.

Future work could explore multi-operator datasets for stylistic diversity, extend the system to dynamic subjects or complex camera motions (e.g., arcs, tracking shots), and incorporate semantic understanding of scene context or composition rules. These additions would expand applicability while preserving artistic intent.

Overall, our results show that LfD offers a robust foundation for creative robotics—bridging intuitive human demonstrations with real-world autonomy in service of cinematic storytelling.

\bibliographystyle{ieeetr}
\bibliography{references.bib}

\begin{thebibliography}{10}

\bibitem{Chen2014}
J.~Chen and P.~Carr, ``{Autonomous camera systems: A survey},'' {\em AAAI Workshop - Technical Report}, vol.~WS-14-06, pp.~18--22, 2014.

\bibitem{Lorimer2024}
P.~Lorimer, J.~Saunders, A.~Hunter, and W.~Li, ``Reinforcement learning of dolly-in filming using a ground-based robot,'' in {\em 2024 IEEE/RSJ International Conference on Intelligent Robots and Systems (IROS)}, pp.~549--556, 2024.

\bibitem{Argall2009}
B.~Argall, {\em Learning Mobile Robot Motion Control from Demonstration and Corrective Feedback}.
\newblock PhD thesis, Carnegie Mellon University, Pittsburgh, PA, March 2009.

\bibitem{Bonatti2020}
R.~Bonatti, Y.~Zhang, S.~Choudhury, W.~Wang, and S.~Scherer, {\em Autonomous Drone Cinematographer: Using Artistic Principles to Create Smooth, Safe, Occlusion-Free Trajectories for Aerial Filming}, pp.~119--129.
\newblock 01 2020.

\bibitem{dang2020enabledronefilmmaker}
Y.~Dang, ``Can we enable the drone to be a filmmaker?,'' 2020.

\bibitem{pygame}
P.~Shinners, ``Pygame.'' \url{http://pygame.org/}, 2011.

\bibitem{coumans2019}
E.~Coumans and Y.~Bai, ``Pybullet, a python module for physics simulation for games, robotics and machine learning.'' \url{http://pybullet.org}, 2016--2023.

\bibitem{brockman2016openai}
G.~Brockman, V.~Cheung, L.~Pettersson, J.~Schneider, J.~Schulman, J.~Tang, and W.~Zaremba, ``Openai gym,'' {\em arXiv preprint arXiv:1606.01540}, 2016.

\bibitem{osa2018algorithmic}
T.~Osa, J.~Pajarinen, G.~Neumann, J.~Bagnell, P.~Abbeel, and J.~Peters, ``An algorithmic perspective on imitation learning,'' {\em Foundations and Trends in Robotics}, vol.~7, pp.~1--179, 11 2018.

\bibitem{gleave2022imitation}
A.~Gleave, M.~Taufeeque, J.~Rocamonde, E.~Jenner, S.~H. Wang, S.~Toyer, M.~Ernestus, N.~Belrose, S.~Emmons, and S.~Russell, ``imitation: Clean imitation learning implementations.'' arXiv:2211.11972v1 [cs.LG], 2022.

\bibitem{schulman2017proximalpolicyoptimizationalgorithms}
J.~Schulman, F.~Wolski, P.~Dhariwal, A.~Radford, and O.~Klimov, ``Proximal policy optimization algorithms,'' 2017.

\bibitem{stable-baselines3}
A.~Raffin, A.~Hill, A.~Gleave, A.~Kanervisto, M.~Ernestus, and N.~Dormann, ``Stable-baselines3: Reliable reinforcement learning implementations,'' {\em Journal of Machine Learning Research}, vol.~22, no.~268, pp.~1--8, 2021.

\bibitem{GAIL2016}
J.~Ho and S.~Ermon, ``Generative adversarial imitation learning,'' in {\em Proceedings of the 30th International Conference on Neural Information Processing Systems}, NIPS'16, (Red Hook, NY, USA), p.~4572–4580, Curran Associates Inc., 2016.

\bibitem{GAIL_in_construction}
R.~Li and Z.~Zou, ``Enhancing construction robot learning for collaborative and long-horizon tasks using generative adversarial imitation learning,'' {\em Advanced Engineering Informatics}, vol.~58, p.~102140, 2023.

\bibitem{GAIL_manipulation}
Y.~Tsurumine, Y.~Cui, K.~Yamazaki, and T.~Matsubara, ``Generative adversarial imitation learning with deep p-network for robotic cloth manipulation,'' in {\em 2019 IEEE-RAS 19th International Conference on Humanoid Robots (Humanoids)}, pp.~274--280, 2019.

\bibitem{kadian2020}
A.~Kadian, J.~Truong, A.~Gokaslan, A.~Clegg, E.~Wijmans, S.~Lee, M.~Savva, S.~Chernova, and D.~Batra, ``Sim2real predictivity: Does evaluation in simulation predict real-world performance,'' {\em IEEE Robotics and Automation Letters}, vol.~PP, pp.~1--1, 08 2020.

\end{thebibliography}

\end{document}